\pdfoutput=1
\documentclass[11pt]{article}

\usepackage{ACL2023}

\usepackage{times}
\usepackage{latexsym}

\usepackage[T1]{fontenc}
\usepackage[utf8]{inputenc}

\usepackage{microtype}

\usepackage{inconsolata}

\usepackage{amsmath}
\usepackage{amssymb}
\usepackage{mathtools}
\usepackage{amsthm}

\usepackage{caption}
\usepackage{subcaption}
\usepackage{acronym}

\usepackage{booktabs}
\usepackage{multirow}

\renewcommand{\[}{\begin{eqnarray}}
\renewcommand{\]}{\end{eqnarray}}

\DeclareMathOperator{\E}{\mathbb{E}}
\DeclareMathOperator{\R}{\mathbb{R}}

\hypersetup{
    colorlinks,
    linkcolor={red!50!black},
    citecolor={blue!50!black},
    urlcolor={blue!80!black}
}

\newcommand{\pmr}[1]{\scriptsize$\pm$#1}

\acrodef{GSLM}{Generative Spoken Language Modeling}

\theoremstyle{plain}
\newtheorem{theorem}{Theorem}[section]

\theoremstyle{definition}
\newtheorem{definition}[theorem]{Definition}

\theoremstyle{remark}

\title{Augmentation Invariant Discrete Representation for \\ Generative Spoken Language Modeling}
\author{
    Itai Gat$^\diamondsuit$, Felix Kreuk$^\diamondsuit$, Tu Anh Nguyen$^\diamondsuit$, Ann Lee$^\diamondsuit$, Jade Copet$^\diamondsuit$, \\ \textbf{Gabriel Synnaeve$^\diamondsuit$, Emmanuel Dupoux$^{\spadesuit, \diamondsuit}$, Yossi Adi$^{\heartsuit,\diamondsuit}$}\\    
    $^\diamondsuit$FAIR Team, Meta AI Research\\
    $^\spadesuit$ENS, INRIA, INSERM, UPEC, PSL Research University\\
    $^\heartsuit$The Hebrew University of Jerusalem\\    
  }
\begin{document}
\maketitle
\begin{abstract}
    \emph{Generative Spoken Language Modeling} research focuses on optimizing speech Language Models (LMs) using raw audio recordings without accessing any textual supervision. Such speech LMs usually operate over discrete units obtained from quantizing internal representations of self-supervised models. Although such units show impressive modeling results, their robustness capabilities have not been extensively investigated. This work focuses on improving the robustness of discrete input representations for generative spoken language modeling. First, we formally define how to measure the robustness of such representations to various signal variations that do not alter the spoken information (e.g., time-stretch). Next, we empirically demonstrate how current state-of-the-art representation models lack robustness to such variations. To overcome this, we propose an effective and efficient method to learn robust discrete speech representation for generative spoken language modeling. The proposed approach is based on applying a set of signal transformations to the speech signal and optimizing the model using an iterative pseudo-labeling scheme. Our method significantly improves over the evaluated baselines when considering encoding and modeling metrics. We additionally evaluate our method on the speech-to-speech translation task, considering Spanish-English and French-English translations, and show the proposed approach outperforms the evaluated baselines.
\end{abstract}

\section{Introduction}

Self-supervised speech models were shown to learn effective representations for various downstream tasks~\citep{hubert, chen2022wavlm, baevski2020wav2vec}. These models were mainly evaluated on discriminative tasks, such as automatic speech recognition, speaker verification, intent classification, etc.~\citep{yang2021superb}. Recently, \citet{on_generative} demonstrated that such self-supervised learning (SSL) representations can be used for Generative Spoken Language Modeling. 

Generative Spoken Language Modeling (GSLM) is the task of learning the acoustic and linguistic characteristics of a language from raw audio. In other words, a discrete representation of the audio signal is being learned. A common practice is to extract continuous representation using an SSL model, then apply vector quantization, usually using the k-means algorithm~\citep{on_generative, kharitonov2021text, borsos2022audiolm}. Then a speech-language model is trained on top of the obtained representation. Finally, a neural vocoder converts the output units to raw audio. As the discrete speech representation often operates over units extracted over relatively short windows (e.g., 20ms), sequences can be long and contain repetitions, e.g., \texttt{10 11 11 11 21 32 32 32 21}. Preliminary studies have found that removing sequential repetitions of units improves performance, hence applying it universally~\citep{on_generative}. For example, a pseudo-text \texttt{10 11 11 11 21 32 32 32 21} becomes \texttt{10 11 21 32 21}. This framework was shown to be effective in modeling multiple levels of the speech utterance, namely prosody, and content~\citep{on_generative, kharitonov2021text, borsos2022audiolm}, speech codec~\citep{polyak2021speech}, speech emotion conversion~\citep{kreuk2021textless}, spoken dialogue~\citep{nguyen2022generative}, and speech-to-speech translation~\citep{lee2021direct, popuri2022enhanced, lee-etal-2022-textless}.

An essential prerequisite for such an audio representation to be used in real-world conditions is robustness to various signal corruptions. Although the aforementioned audio representation models have shown effectiveness in many tasks, they were mainly evaluated on academic benchmarks.

\begin{figure*}
    \centering\vspace{-0.2cm}
    \includegraphics[width=0.9\textwidth]{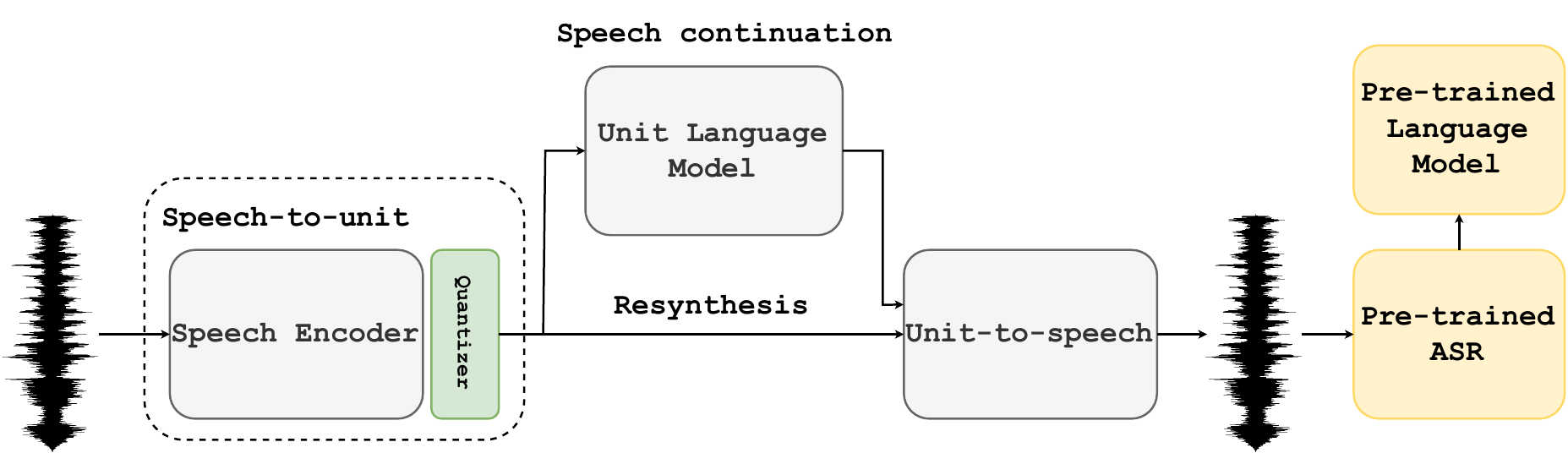}
    \caption{Generative Spoken Language Modeling is composed of three components: (i) Speech-to-unit, (ii) Unit language model, and (iii) Unit-to-speech. Pre-trained ASR and language models are used for evaluation.}
    \label{fig:gslm}
    \vspace{-0.2cm}
\end{figure*}

In this work, we evaluate current state-of-the-art self-supervised speech representation models on what are arguably the most basic signal variations, namely time-stretch, pitch-shift, additive-noise, and reverberation. Our premise is that while these variations modify the signal, its' underlying content remains the same, especially under the units repetition removal process. Therefore, a robust representation should be affected by such variations to a minimal extent.

As a first step, we propose a set of metrics for evaluating the model's robustness. Then, we point to the lack of robustness of these models with respect to the aforementioned variations. Next, we design a simple and effective method for learning robust discrete representation on top of any speech SSL model. We demonstrate how such a method greatly improves robustness. Then, we empirically show that performance improves on several tasks for various SSL models. Specifically, we evaluate the newly proposed speech encoders when considering zero-shot evaluation tasks considering encoding and modeling, i.e., ABX, sWUGGY, and sBLIMP~\citep{nguyen2020zero}, together with a high-level downstream task in the form of speech-to-speech translation.

\section{Background}
The general \ac{GSLM} pipeline is comprised of three main modules: (i) Speech-to-unit, (ii) Unit language model, and (iii) Unit-to-speech, where each of these modules is trained separately. Speech resynthesis can be achieved while ignoring the language model and directly feeding the quantized units into the unit-to-speech module~\citep{polyak2021speech} (See Figure~\ref{fig:gslm} for a visual description). In the following paragraphs, we give detailed background for each of the three components mentioned above, including the standard evaluation methods.  

\paragraph{Speech-to-unit} module encodes the raw speech signal into a discrete representation. The common approach is first to encode the speech into a continuous representation and then quantize the representation to achieve a sequence of discrete units~\citep{on_generative, polyak2021speech, popuri2022enhanced, lee2021direct, kharitonov2021text, kreuk2021textless, kharitonov2022textless, nguyen2022generative, borsos2022audiolm, tjandra2019vqvae, tjandra2020transformer}. 

Formally, denote the domain of audio samples by $\mathcal{X} \subset \mathbb{R}$. The representation for a raw signal is therefore a sequence of samples $x = (x_1,\ldots, x_T)$, where  $x_t\in\mathcal{X}$ for all $1\leq t \leq T$. 

Consider an encoder network, $f$, that gets as input the speech utterance and outputs a sequence of spectral representations sampled at a low frequency as follows $f(x) = (v_1, \dots, v_{T'})$. Note that we do not assume anything about the structure of the encoder network $f$. \citet{on_generative}, evaluated several speech encoders, namely, Mel-spectrogram, Contrastive Predictive Coding~\citep[CPC]{oord2018representation}, wav2vec2~\citep{baevski2020wav2vec}, and HuBERT~\citep{hubert}. 

Since the representations learned by such models are usually continuous, a k-means algorithm is applied over the models' outputs to generate discrete units, denoted as $z = (z_1,\ldots,z_{T'})$. Each element $z_i$ in $z$ is a positive integer, $z_i\in\{1,..,K\}$ for $1\le i \le T'$, where $K$ is the number of discrete units. We denote the quantization model with $E$.

\paragraph{Unit Language Model}  is trained on the extracted discrete units, $z$. Such a language model learns a probability distribution of the learned unit sequences, which enables direct modeling of speech data without textual supervision. 

The language model can be used to generate speech conditionally or unconditionally, replicating what toddlers achieve before learning to read. Moreover, such a modeling framework allows for capturing and modeling prosodic features~\citep{kharitonov2021text}, as well as speaker identity~\citep{borsos2022audiolm}, or even natural dialogues~\citep{nguyen2022generative}. This is in contrast to using textual features, as they do not encode such information.

\paragraph{Unit-to-speech} module converts the speech discrete units to a raw waveform. \citet{on_generative} used a Tacotron2.0~\citep{tac} based model followed by WaveGlow~\citep{prenger2019waveglow} vocoder. Later, \citet{polyak2021speech} proposed a unit-based vocoder based on the HiFi-GAN architecture to convert units to speech directly. Such a paradigm seems to provide high-quality generations with better efficiency as it uses only one model rather than two. \citet{kreuk2021textless} and \citet{lee2021direct} additionally improved the unit based vocoder to include emotional tokens for speech emotion conversion tasks, and duration modeling for direct speech-to-speech translation. 

\paragraph{Zero-shot Evaluation.} Evaluating such a complex pipeline comprised of several components is a challenging task. \citet{on_generative} proposed a set of zero-shot evaluation tasks aiming for each of the modules. Overall the proposed tasks can be divided into four main groups: (i) acoustic encoding using ABX, bitrat, (ii) language encoding using sWUGGY, sBLIMP~\citep{nguyen2020zero, on_generative}, (iii) resynthesis using Phoneme/Word Error Rate; (iv) speech generation using VERT~\citep{on_generative}, Meaningfulness Mean Opinion Score.

\begin{figure*}
     \centering\vspace{-0.2cm}
     \begin{subfigure}[b]{0.24\textwidth}
         \centering         
         \includegraphics[width=\textwidth]{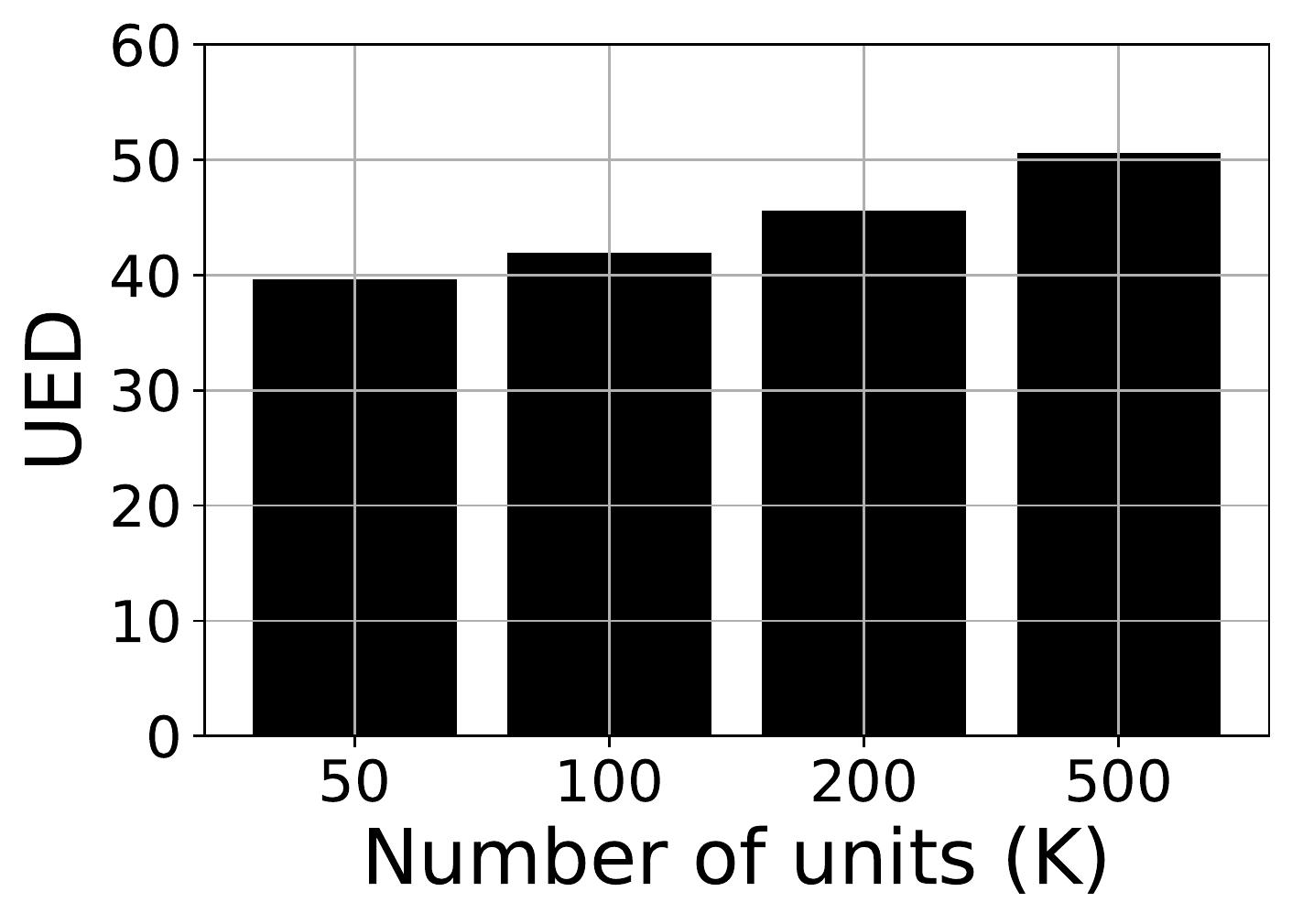}
         \caption{Time stretch}
     \end{subfigure}
    %  \hfill
     \begin{subfigure}[b]{0.24\textwidth}
         \centering
         \includegraphics[width=\textwidth]{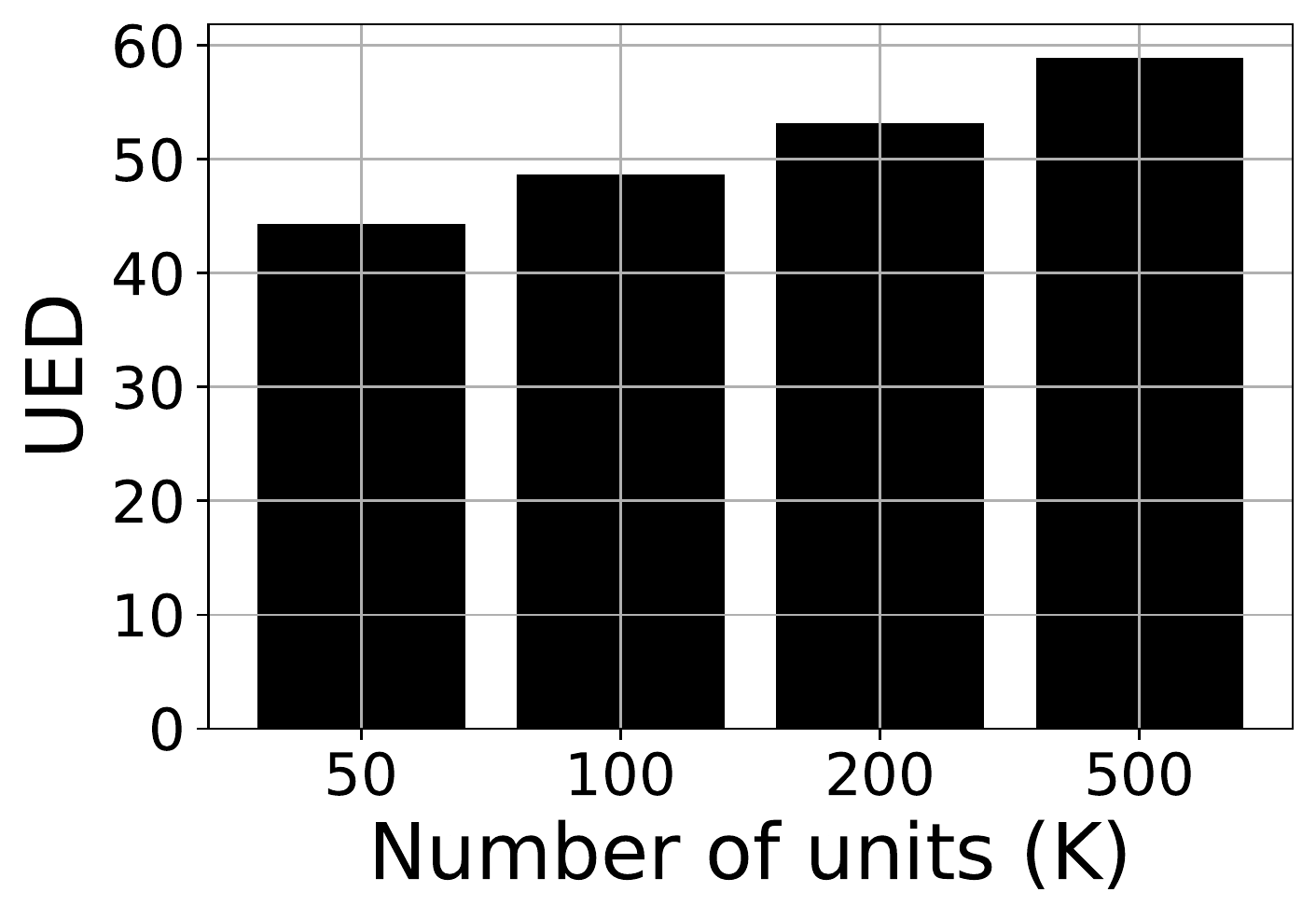}
         \caption{Pitch shift}
     \end{subfigure}
     \begin{subfigure}[b]{0.24\textwidth}
         \centering
         \includegraphics[width=\textwidth]{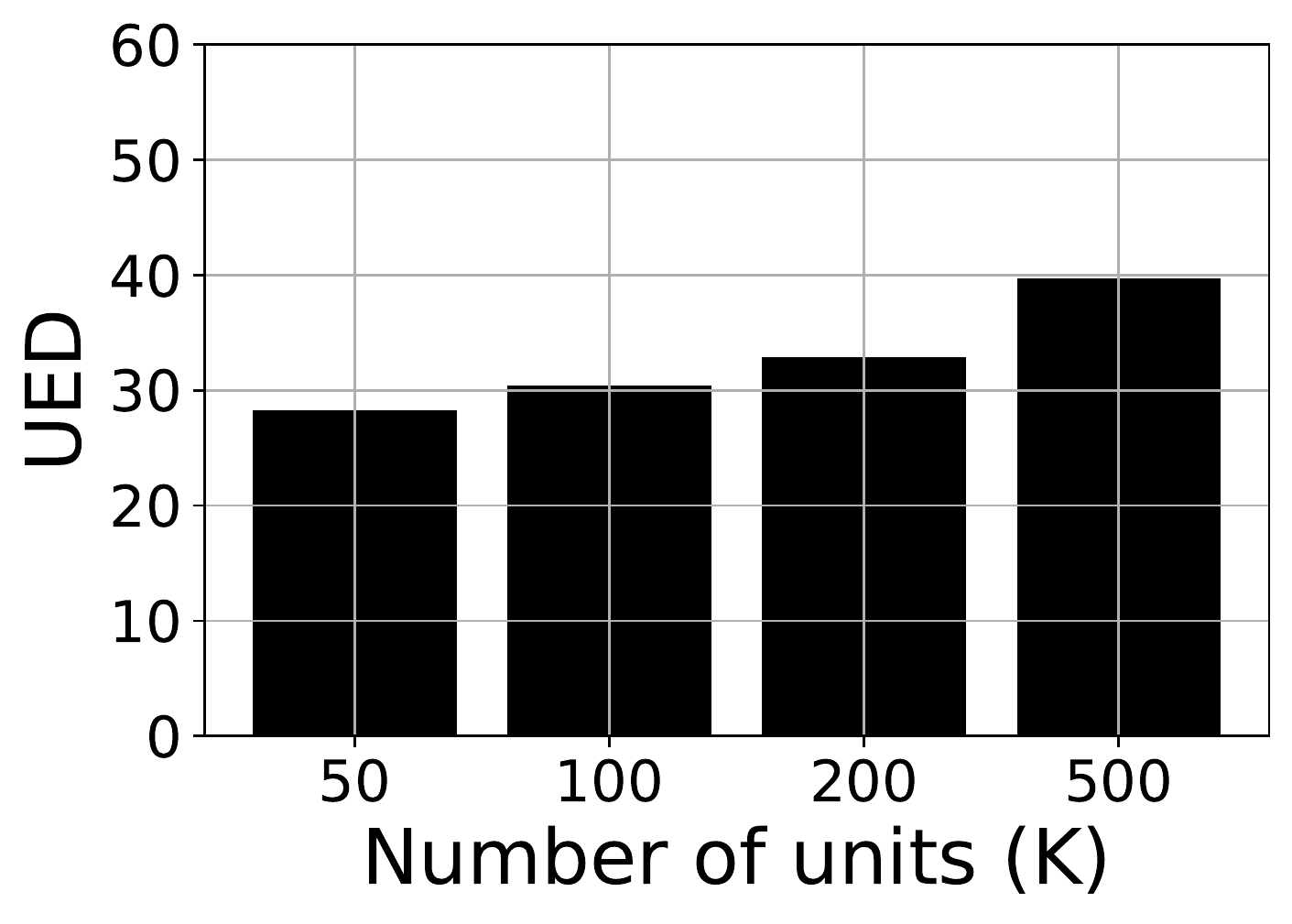}
         \caption{Reverberation}
     \end{subfigure}
     \begin{subfigure}[b]{0.24\textwidth}
         \centering
         \includegraphics[width=\textwidth]{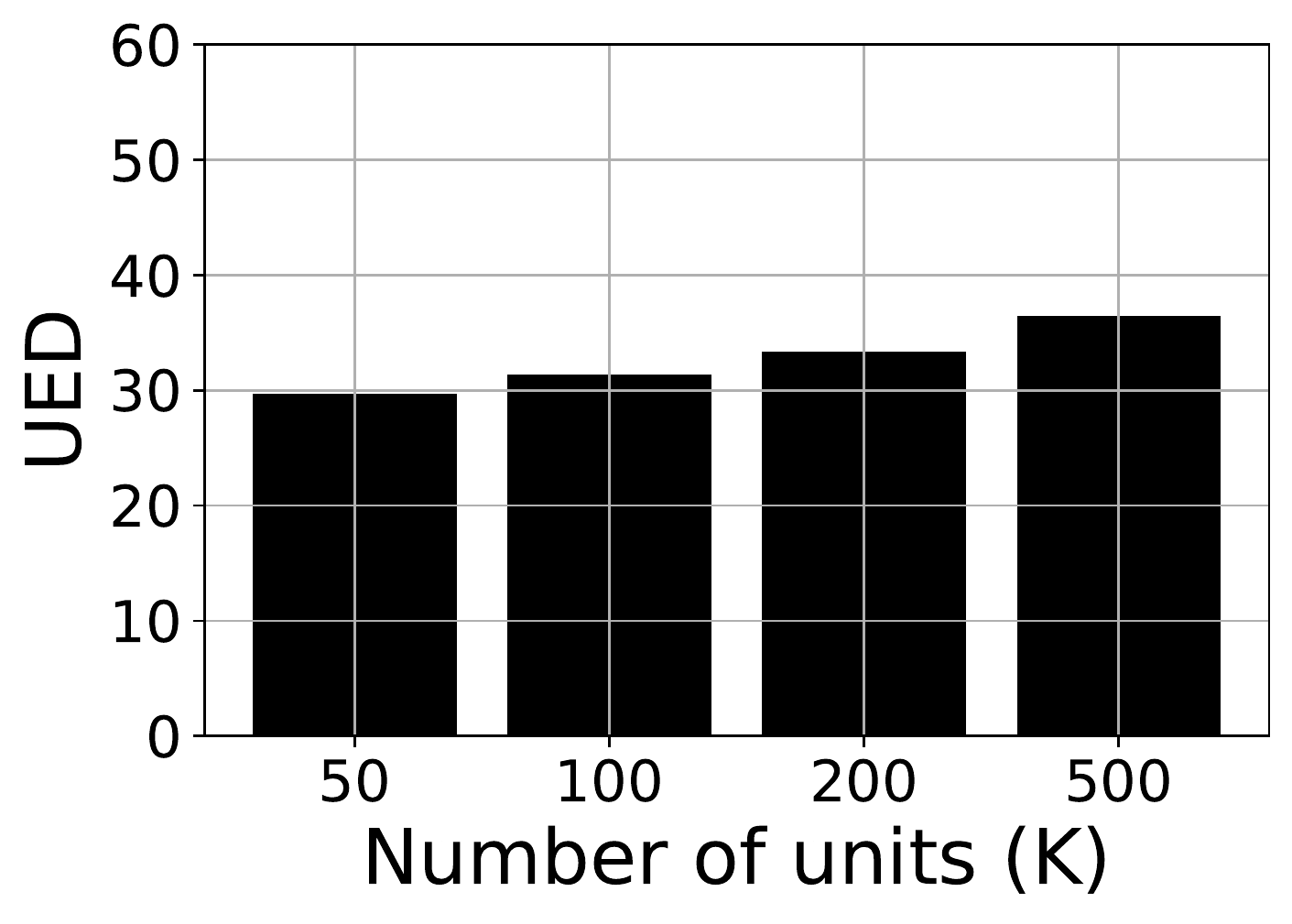}
         \caption{Noise}
     \end{subfigure}
     \caption{UED scores for various augmentations and number of clusters. We note that the UED is relatively high (the distance is normalized). We also note that the UED monotonically increases with the number of units used. We multiply the scores by a hundred.}
     \label{fig:kmeans_ued}
     \vspace{-0.2cm}
\end{figure*}

\section{Robustness of Speech-to-Unit Models}\label{sec:robust}

The first step toward developing an effective spoken language model is to develop a robust representation. The focus of a robust representation should be on the spoken information rather than unrelated signals, such as prosodic features in the form on duration and F0, background noise, or reverberations. In the following section, we propose a metric for quantifying the degree to which augmentations change the resulting encoding.

\subsection{Unit Edit Distance}

A spoken language model is built on top of a discrete representation of a continuous encoder. We examine the robustness of the discrete space to augmentations that do not change the spoken content. Therefore, we are interested in a sequential distance metric between two discrete representations. It is essential to note that augmentations can alter the spatial dimension of the signal. For example, stretching a signal results in more frames, yielding a longer representation sequence. Similar phenomenon will happen when convolving with different room impulse response to simulate reverberation. Hence, the metric should be able to measure the distance between two sequences of different lengths. Ideally, it will consider the number of deletions, insertions, and substitutions that occur due to augmenting the input data. For this purpose, we find the Levenshtein distance a good fit~\citep{levenshtein1966binary}. The Levenshtein distance measures the minimum changes one should make to modify one sequence to another. It has two essential properties: the first is that the score is non-negative, and when the sequences are equal, the metric equals zero. The second property is that the maximum value it can get equals the longer sequence length between the two sequences. We provide a detailed explanation of the Levenshtein distance in the Appendix material.

We aggregate the distance values over the evaluation set while considering the sequence length. This is desirable since we want to normalize scores for sequences in different lengths, and the Levenshtein distance's maximum value is the original sequence's length. Another essential property of a spatial metric is repetitions. Consider time stretch as an example, it changes the number of the input frames, but one would expect the deduplicated quantized signal to be the same as before the augmentation. Hypothetically, one can maximize the score by stretching the signal infinitely. To eliminate such dependencies, we compute the score on a deduplicated quantized representation. Formally, our final metric is:

\begin{definition}[Unit Edit Distance]
    Given a continuous encoder $f: \R^{T}\rightarrow\R^{T'}$, a quantizer $E: \R^{T'}\rightarrow\{1,..,K\}^{T'}$, and an input augmentation $g: \R^{T'}\rightarrow\R^{\widehat{T'}}$. The deduplicated unit edit distance $\text{UED}_{\mathcal{D}}(E, f, g)$ on the evaluation set $\mathcal{D}$ is:
    \[
         \sum_{x\in \mathcal{D}} \frac{1}{T'_{x}} \text{LEV}\left((E\circ f)(x), (E\circ f \circ g)(x)\right),
    \]
    where $T'_{x}$ is the number of frames of a sample $x$.
\end{definition}
Ideally, a perfect spoken language quantizer obtains a zero distance after deduplication. Next, we study state-of-the-art spoken language representations using our proposed metric in different settings.

\subsection{Evaluation}

In the following, we study current state-of-the-art representations for generative spoken language modeling using the proposed metric. The current popular quantization technique is a k-means model trained on top of a pre-trained encoder~\citep{on_generative}. In our evaluation setup, we use a different number of clusters and encoder architectures. Our ablation study include quantizers with $50$, $100$, $200$, and $500$ clusters. We further investigate our metric on top of HuBERT~\citep{hubert}, wav2vec2~\citep{baevski2020wav2vec}, and WavLM~\citep{chen2022wavlm}. For readability, throughout the paper, we report results for the HuBERT model while leaving the rest of the results in the Appendix material.

\subsubsection{Augmentations}\label{sec:aug}

This work focus on four simple signal modifications which mimic real-world signal variations: 

\paragraph{Time stretch.} We use the Phase Vocoder method~\citep{karrer2006phavorit} to stretch or shrink the time domain signal with a rate of $\tau$ without changing the pitch. For example, $\tau=1.2$ speeds up the signal by $20\%$. In this work, for each sample, we sample uniformly a value in the range $[0.8, 1.2]$.

\paragraph{Pitch shift.} We change the original pitch of the speech signal by a given number of semitones using the resampling method over the time-stretched signal~\citep{karrer2006phavorit}. In this paper, we shift the pitch by up to four semitones.
    
\paragraph{Reverberation.} We follow a similar setting of~\citet{chazan2021single}, in which we consider an Acoustic Transfer Function (ATF) to be simulated using the pyroomacoustics~\citep{scheibler2018pyroomacoustics} audio room simulations package. We randomly sample room dimensions, microphone location, and source location, then convolve the ATF with the speech signal. 

\paragraph{Noise injection.} We mix a given speech signal with non-stationary additive noise, using a randomly sampled Signal-to-Noise Ratio (SNR) in the range of $[5, 15]$. Background noises are sampled from the Deep Noise Suppression (DNS) challenge~\citep{reddy2020interspeech} which includes a diverse set of noise types from AudioSet~\citep{gemmeke2017audio}, Freesound,~\footnote{\url{https://freesound.org/}} and Demand~\citep{thiemann_joachim_2013_1227121}.

\subsubsection{Results}

In Figure~\ref{fig:kmeans_ued}, we use our metric to study the robustness of k-means trained on top of HuBERT with various augmentations and values of $K$. This evaluation points to the lack of robustness of the current state-of-the-art representation of simple, non-spoken augmentations. For example, for time stretch augmentation, the UED score is between $39$ and $51$. Considering that UED is computed on deduplicated signals, those numbers are high. Moreover, this number increases as a function of $K$. The high numbers and the monotonicity of the UED as a function of $K$ are consistent for all values of $K$, augmentations, and models we experimented with (HuBERT, wav2vec2, and WavLM). Next, we propose a method that improves the robustness of such representations.

\begin{figure*}
    \centering\vspace{-0.2cm}
    \includegraphics[width=0.9\textwidth]{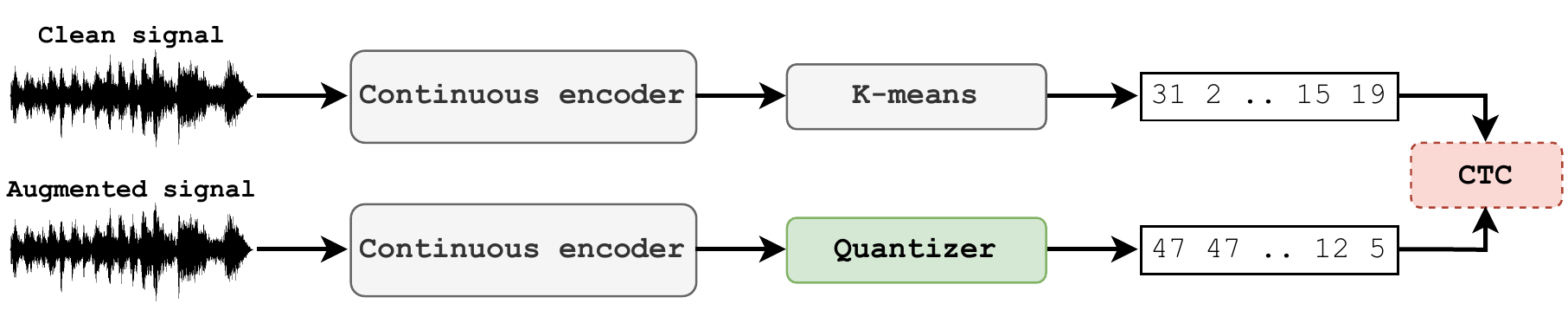}
    \caption{Illustration of our method: We forward a clean signal through an encoder followed by a pre-trained quantizer (k-means). Next, we forward an augmented signal through the same encoder, followed by a new quantizer (green). The CTC loss between the deduplicated output of the clean signal and the output of the augmented signal is used to learn the parameters of the new quantizer. In the iterative approach, post the convergence of the learned quantizer $E_0$, we freeze it and learn a new quantizer $E_1$ that distills information from $E_0$.}
    \vspace{-0.2cm}
\end{figure*}

\section{Pseudo-labeling for Robust Discrete Representation}

Our findings in Section~\ref{sec:robust} suggest that current state-of-the-art representations may be too sensitive to augmentations that do not alter spoken information. Preliminary robustness research focused primarily on noise augmentation. This is convenient since the signal length is not affected by such augmentations. In practice, real-world augmentations may modify the signal length. In order to work with various types of augmentations, we must align the original and augmented sequences. The following section presents a pseudo-labeling, alignment-based approach to learning a robust quantizer.

\subsection{Pseudo-labeling}\label{sec:pseudo_iter}

The GSLM encoding framework comprises a raw audio signal forwarded through an encoder, then a quantizer. The quantizer is learned on top of a trained encoder, e.g., k-means trained on each embedding vector extracted from HuBERT. 

As discussed above, we do not want to limit the robustness process to a family of augmentations that do not change the signal's length. To align and use augmentations that may modify the signal's length, we use the Connectionist Temporal Classification (CTC) loss~\citep{graves2006connectionist}. The CTC operation computes the probability of an alignment based on the predicted and target sequences. Finally, the CTC loss considers the negative log-likelihood produced by the CTC operation.

% We forward a clean signal through an encoder $f$ followed by a pre-trained quantizer $E_0$. Parallelly, we forward an augmented signal through the same encoder $f$ and train a non-linear multi-layer-perceptron $E_1$. Using the CTC loss, which accounts for the alignment between the outputs, we learn the parameters of $E_1$. Formally, the probability given by the CTC loss for a single data point $x$ follows
% \[
%     \ell(E_0, E_1, x, g) &\triangleq& - p\left((E_0\circ f)(x) | (E_1\circ f\circ g)(x)\right)
%     \\ &=& - \sum_{\mathcal{A}\in \mathcal{A}_x} \prod_{t=1}^{r} p_t(a_t | (E_1\circ f\circ g)(x)),
% \]
We forward a clean signal through an encoder $f$ followed by a pre-trained quantizer $E_0$. Parallelly, we forward an augmented signal through the same encoder $f$ and train a non-linear multi-layer-perceptron $E_1$. Using the CTC loss, which accounts for the alignment between the outputs, we learn the parameters of $E_1$. Formally, the probability given by the CTC loss $\ell(E_0, E_1, x, g)$ for a single data point $x$ follows
\[
     - p\left((E_0\circ f)(x) | (E_1\circ f\circ g)(x)\right),
\]
which can be decomposed to a sum over the set of all alignments $\mathcal{A}_x$
\[
    - \sum_{\mathcal{A}\in \mathcal{A}_x} \prod_{t=1}^{r} p_t(a_t | (E_1\circ f\circ g)(x)).
\]

Finally, for a training set $\mathcal{D}$, a set of augmentations $\mathcal{G}$, a pre-trained quantizer $E_0$, and a learned quantizer $E_1$, our loss function is as follows:
\[
    \mathcal{L}_{\mathcal{D}}(E_0, E_1, \mathcal{G}) \triangleq\E_{x\sim \mathcal D, g\sim \mathcal{U}(\mathcal{G})} \left[\ell(E_0, E_1, x, g)\right].\nonumber
\]

Note that the alignment between the predicted and target sequences is many-to-one. Thus, one or more output units can be aligned to a single target unit. Hence, to work with augmentations that stretch the signal, we are required to deduplicate the target sequence. Intuitively, this process distills quantization knowledge from the pre-trained quantizer into the new quantizer while injecting $E_1$ knowledge about the contextual similarity between the original and augmented signals. 

A significant advantage of our method is that it is highly efficient. Our method requires training only a relatively small amount of parameters. In contrast to previous methods that train HuBERT from scratch, which takes up to seven days on $32$ GPUs, our method converges in a few hours on a single GPU. In fact, our experiments show that learning the parameters of the encoder performs worse than freezing them. While the UED is boosted, but the ABX are negatively affected. The freezing of the upstream model thus serves as a regularizer.

\begin{table*}[t!]\centering\small\vspace{-0.2cm}
\begin{tabular}{llcccc}
    \toprule
    \multirow{2}[3]{*}{\# units} &\multirow{2}[3]{*}{Method} & \multicolumn{4}{c}{Augmentation} \\
    \cmidrule(lr){3-6}
     &                                    & Time                    & Pitch shift             & Reverberation           & Noise \\
    \midrule
    \multirow{3}{*}{50} &k-means          & 39.61\pmr{0.37}         & 44.33\pmr{0.92}         & 28.25\pmr{0.61}         & 29.74 \pmr{0.31} \\
                        &Ours             & 27.91\pmr{0.42}         & 30.74\pmr{0.71}         & 20.16\pmr{0.60}         & 25.33\pmr{0.36} \\
                        &Ours (Iterative) & \textbf{26.89}\pmr{0.33}& \textbf{30.22}\pmr{0.79}& \textbf{19.89}\pmr{0.54}& \textbf{24.67}\pmr{0.29} \\
    \midrule
    \multirow{3}{*}{100} &k-means         & 41.97\pmr{0.42}         & 48.68\pmr{0.96}         & 30.42\pmr{0.69}         & 31.38\pmr{0.33}\\
                         &Ours            & 31.05\pmr{0.39}         & 34.77\pmr{0.92}         & 22.21\pmr{0.63}         & 28.05\pmr{0.31}\\
                         &Ours (Iterative)& \textbf{29.72}\pmr{0.41}& \textbf{32.84}\pmr{0.91}& \textbf{21.31}\pmr{0.71}& \textbf{25.06}\pmr{0.31}\\
    \midrule
    \multirow{3}{*}{200} &k-means         & 45.59\pmr{0.39}         & 53.14\pmr{1.01}         & 32.89\pmr{0.72}         & 33.34 \pmr{0.38}\\
                         &Ours            & 34.40\pmr{0.46}         & 38.51\pmr{1.09}         & 24.10\pmr{0.66}         & 30.19\pmr{0.37}\\
                         &Ours (Iterative)& \textbf{32.99}\pmr{0.42}& \textbf{36.45}\pmr{1.03}& \textbf{22.94}\pmr{0.67}& \textbf{26.76} \pmr{0.31} \\
    \midrule
    \multirow{3}{*}{500} &k-means         & 50.60\pmr{0.42}         & 58.92\pmr{0.98}         & 39.71\pmr{0.81}         & 36.47\pmr{0.44}\\
                         &Ours            & 38.04\pmr{0.44}         & 43.48\pmr{1.03}         & 28.43\pmr{0.73}         & 29.99\pmr{0.45}\\
                         &Ours (Iterative)& \textbf{36.50}\pmr{0.49}& \textbf{40.82}\pmr{1.02}& \textbf{25.78}\pmr{0.74}& \textbf{27.51}\pmr{0.49}\\
    \bottomrule
\end{tabular}
\caption{Unit edit distance study: Using our metric, we assess the robustness of various quantization methods on top of a HuBERT representation. This study uses four different augmentations: time stretching, pitch shifting, reverberation, and noise injection. The non-iterative (Section~\ref{sec:pseudo_iter}) and iterative (Section~\ref{sec:pseudo_non_iter}) methods significantly and consistently improve the robustness of k-means. Pseudo-labeling accounts for most of the improvement. By applying our method iteratively, we can improve it further. For readability, we multiply the scores by a hundred.}
\label{table:ued}
\vspace{-0.2cm}
\end{table*}

\subsection{Iterative Pseudo-labeling}\label{sec:pseudo_non_iter}

In the previous section, we presented a pseudo-labeling approach that relies on a converged quantizer $E_0$, e.g., k-means on top of HuBERT. This raises the question of whether it is possible to enhance the robustness of the learned quantizer $E_1$ by iteratively replacing the pre-trained quantizer with the converged quantizer and learning another MLP on top of it. It turns out that such a process can further improve the robustness.

The iterative process begins with a pre-trained quantizer $E_0$, then, as in Section~\ref{sec:pseudo_iter} we learn a robust quantizer $E_1$. Upon $E_1$ convergence, we replace $E_0$ with $E_1$ and use it as the pre-trained quantizer. Then, we learn a new MLP $E_2$ on top of the converged $E_1$. We repeat this process $K$ times. This process needs more careful training. We note that it is essential to replace the quantizers only post-convergence.

\section{Experiments}

In the following, we assess the efficacy of our method using state-of-the-art self-supervised representations and popular discriminative and generative evaluation tasks. It is important to note that a single metric cannot tell the whole story. For example, similarly to perplexity, all representations can be assigned to the same cluster, which achieves a perfect unit edit distance but a poor representation. We first examine our proposed method using the unit edit distance along with other discriminative and generative performance metrics. Then, we show that our method improves downstream tasks.

In Section~\ref{sec:exp_ued} we use our proposed metric from Section~\ref{sec:robust} to analyze the robustness of our method. In Section~\ref{sec:exp_abx} we study the discriminative capabilities of our method using the ABX test~\citep{abx}. Then, we evaluate our methods using generative zero-shot evaluation tasks such as sWUGGY and sBLIMP~\citep{nguyen2020zero, on_generative}. Finally, we demonstrate the effect of using our robust quantizer's units in speech-to-speech translation.

\paragraph{Experimental Setup.} We study our method using the base versions of HuBERT, wav2vec2, and WavLM. For readability, we report results for HuBERT in the main paper. The results for wav2vec2 and WavLM are in Appendix~\ref{app:results}. To match the current k-means training set, we use the Librispeech-100h to learn our quantizer~\citep{librispeech}. We analyze our metric using the `clean' and `other' development sets from Librispeech. A detailed setup is provided in Appendix~\ref{app:exp_setup}.

% \vspace{-0.1cm}
% \subsection{Experimental Setup}\label{sec:exp_setup}

% In the following, we detail in short the experimental setup. A detailed is provided in Appendix~\ref{app:exp_setup}.

% \noindent\textbf{Models.} We study our method using the base versions of HuBERT, wav2vec2, and WavLM. Similar to prior work, for HuBERT and WavLM, we use the ninth and sixth layers for wav2vec2. For readability, we report results for HuBERT in the main paper. The results for wav2vec2 and WavLM are presented in Appendix~\ref{app:results}.

% \noindent\textbf{Datasets.} To match the current k-means training set, we use the Librispeech-100h to learn our quantizer~\citep{librispeech}. We analyze our metric using the `clean' and `other' development sets from Librispeech. The augmentations in all setups include time stretch, pitch shift, reverberation, and noise injection (exact parameters are detailed in Section~\ref{sec:aug}).

\begin{table*}[t!]\centering\small\vspace{-0.2cm}
\begin{tabular}{llccccccc}
    \toprule
    \multirow{2}[3]{*}{\# units} &\multirow{2}[3]{*}{Method} & \multicolumn{2}{c}{ABX (clean) $\downarrow$} & \multicolumn{2}{c}{ABX (other)$\downarrow$}& \multirow{2}[3]{*}{sWUGGY $\uparrow$} & \multirow{2}[3]{*}{sBLIMP $\uparrow$} \\
    \cmidrule(lr){3-4}\cmidrule(lr){5-6}
     &&Within & Across & Within & Across&& \\
    \midrule
    \multirow{3}{*}{50} &k-means& 7.52 & 8.90 & 9.84 & 13.5 & 66.12 & 54.91 \\
    &Ours& 6.76 &7.72 & \textbf{9.03} & \textbf{11.78} &\textbf{67.59}& 55.76\\
    &Ours (Iterative)& \textbf{6.63} & \textbf{7.55} & 9.53 & 12.14 &67.42& \textbf{57.04}\\
    \midrule
    \multirow{3}{*}{100} &k-means& 6.37 & 7.72 & 8.4 & 12.29 &67.70&56.16\\
    &Ours& 5.50 & \textbf{6.21} & \textbf{7.24} & \textbf{10.11}&67.79&\textbf{57.01} \\
    &Ours (Iterative)& \textbf{5.39} & \textbf{6.22} & 7.46 & 10.20&\textbf{68.20}&56.99 \\
    \midrule
    \multirow{3}{*}{200} &k-means& 5.99 & 7.14 & 8.23 & 11.51 &66.51& 54.64\\
    &Ours& 5.29 & \textbf{6.01} & 7.22 & 9.78 &70.51& 56.19\\
    &Ours (Iterative)& \textbf{5.19} & \textbf{6.00} & \textbf{7.18} & \textbf{9.70} &\textbf{70.68}& \textbf{56.26}\\
    \midrule
    \multirow{3}{*}{500} &k-means& 5.98 & 6.98 & 7.89 & 11.43 &66.92& 55.97\\
    &Ours& 5.16 & 6.03 & 7.06 & 9.76 &\textbf{70.13}& 55.19\\
    &Ours (Iterative)& \textbf{4.96} & \textbf{5.73} & \textbf{6.93} & \textbf{9.63} &69.33& \textbf{56.93}\\
    \bottomrule
\end{tabular}
\caption{Zero-shot discriminative and generative evaluation tasks: We evaluate the ABX score on the `clean' and `other' development sets from Librispeech. Our method improves the scores scores in all setups.}
\vspace{-0.3cm}
\label{table:metrics}
\end{table*}

\subsection{Analysis}\label{sec:exp_ued}

In Section~\ref{sec:robust}, we presented an evaluation metric that assesses the robustness of a quantized speech representation to augmentations. The metric is insensitive to changes in the length of the signal. Using it, we investigated the current state-of-the-art representations. In the following, we study our robust quantization method.

Table~\ref{table:ued} presents the unit edit distance metric using our robustness method with and without the iterative approach. Compared with the k-means method, which is currently in use, our non-iterative method consistently outperforms it by a large margin (relative improvement of at least 30\%). We also note that different augmentations affect the representation differently. Our iterative method provides a slight but consistent improvement over the non-iterative method. It is noticeable that the UED is increasing (i.e., worse performing) with the number of units used.

\subsection{Zero-shot Evaluation}\label{sec:exp_abx}
We evaluate the proposed method using the standard GSLM setup, i.e., ABX, sWUGGY, sBLIMP. The ABX task examines the discriminative phonetic abilities of the representation. ~\citet{versteegh2015zero} show that the ABX result is a good proxy to signal content (i.e., Phoneme Error Rate). The input to the ABX is a pair of words with a phoneme modification and a reference word containing the same phoneme as one of the pair's words. Next, the ABX measures the distance of the test phoneme representation to both the correct and incorrect representations. Finally, the distance between the test and the correct representation is expected to be lower than the distance to the incorrect representation. The ABX task is conducted in two setups: `within' and `across.' `Within' is evaluated on input data from the same speaker, while `across' is evaluated on input data from different speakers.

Table~\ref{table:metrics} shows the ABX results for both Librispeech `clean' and `other'. In our experiments, we found that the ABX score consistently and significantly improved on all the setups we tested. In this case, the iterative approach improves more than the non-iterative one, but the improvement is inconsistent. For a small number of units and the `other' split, the ABX score is lower than the iterative model's score. Note that the `other' split is challenging as it is characterized by recordings that contain background noise and various accents.

The spot-the-word task (sWUGGY) requires detecting the real word from a pair of short utterances such as `brick' vs. `blick.' The detection is done by comparing the probabilities given by a language model for each word. This allows comparing representations by training language models on top of them. Differently, the acceptability judgment test (sBLIMP) requires detecting the syntactically correct sentence from a pair of sentences, one of which is syntactically correct and the other is wrong. The detection is based on the perplexity of the language model. As presented in Table~\ref{table:metrics}, our method enables improvement in all the investigated setups for both the spot-the-word and acceptability judgment tests. This is especially noticeable for a larger number of units. For instance, when considering $200$ or $500$ units, the absolute improvement of the sWUGGY score is $4.17$ and $3.21$, respectively. 

\subsection{Speech-to-speech Translation}\label{sec:s2s}

Lastly, we evaluate the proposed method considering the speech-to-speech translation task. To better assess the effectiveness of the proposed robust discrete representation we follow the same setup as in \citet{lee-etal-2022-textless} while changing the discrete speech representation only. 

\citet{lee-etal-2022-textless} propose a textless speech-to-speech translation method by forwarding a source speech signal and predicting its target's discrete representation. The authors use a k-means model trained on top of a multilingual HuBERT (mHuBERT) for speech representation. Additionally, the authors show that solving an auxiliary task enhances performance. We investigate the impact of using our robust quantizer as an alternative to the k-means used by \citet{lee-etal-2022-textless}. Differently, we use HuBERT (instead of mHuBERT). Besides that, we follow the same setup in terms of model, computation resources, and data. To evaluate the quality of the translation the sentence BLEU score (SacreBLEU)~\citep{post-2018-call} was used.

Table~\ref{tab:s2st} presents the results for the Spanish-English and French-English setups on the Europarl-ST development and test sets~\citep{europarl}. It also shows the original result from~\citet{lee-etal-2022-textless}. The proposed method improves over ~\citet{lee-etal-2022-textless} under all the evaluated setups. Note, these results are especially interesting as the proposed method was trained on significantly less data (ours was trained on 1k hours while \citet{lee-etal-2022-textless} was trained on 100k hours).

\begin{table}[t!]
\centering\small
\begin{tabular}{lllccccccc}
    \toprule
    &\# units &Method & S-E & F-E \\
    \midrule
    \multirow{3}{*}{Dev}&500 & Robust & 17.3 & 16.4 \\
    \cmidrule{2-5}
    &1000 & k-means & 15.4  & 16.0\\
    \cmidrule{2-5}
    &1000 & Robust & \textbf{18.2} & \textbf{17.5} \\
    \midrule
    \midrule
    \multirow{3}{*}{Test}&500 &Robust & 14.4 & 15.75 \\
    \cmidrule{2-5}
    &1000 &k-means& 13.1  & 15.4 \\
    \cmidrule{2-5}
    &1000 &Robust & \textbf{15.9} & \textbf{17.1} \\
    \bottomrule
\end{tabular}
\caption{Speech-to-Speech Translation results: We report BLEU scores for the proposed method (Robust) and compare it against the k-means used in~\citet{lee-etal-2022-textless}. We report both development and test sets results for Spanish(S)-English(E) and French(F)-English(E).}
\label{tab:s2st}
\vspace{-0.2cm}
\end{table}

\section{Related work}

This work investigates the robustness of self-supervised representations for language modeling. This is related to the advancements in speech self-supervised learning, their robustness, and modern generative spoken language modeling. In the following, we review all three areas.

\paragraph{Self-supervised Learning.} The field of deep learning research has significantly benefited from self-supervised learning. Commonly, it involves encoding the input data and performing a task that enforces the representation to learn contextual embeddings. Speech self-supervised learning can be divided into two lines of research. 

The first is discriminative, \citet{oord2018representation} introduced Contrastive Predictive Coding (CPC), which trains a convolutional encoder and a predictor for future embeddings of the encoder using a contrastive loss. On top of it,~\citet{kharitonov2021data} propose to use time domain augmentations to improve the CPC model further. Wav2vec2~\citep{schneider2019wav2vec} suggest using a contrastive loss that requires distinguishing between true and false future audio samples. Later, wav2vec2~\citep{baevski2020wav2vec} learn quantized units using Gumbel softmax and predict masked spans of the latent speech representation. HuBERT~\citep{hubert} employ a frame-based masked prediction task. First, it quantizes input frames and then predicts masked frames. 

The second line of work is generative. An early generative self-supervised work is Autoregresstive Predictive Coding~\citep{chung2019unsupervised}, which predicts the spectrum of a future frame. Later,~\citet{liu2020mockingjay} introduced Mockingjay, which learns its representation by predicting non-causal context. TERA~\citep{liu2021tera} alters time, frequency, and magnitude. Then it is required to reconstruct acoustic frames from altered versions.

\paragraph{Robustness.} A desired property of a spoken language representation is robustness to augmentations that do not change the spoken information. The spoken information should not differ significantly when male and female speakers say the same content. There is an interesting trade-off between training a robust representation and the quality of the input data. It is possible, for example, to use the same speaker for all data points in the training set. The model would not be able to learn any speaker bias, but this constraint prevents scaling. 

Recently, the robustness of self-supervised speech representations has gained attention from the community. WavLM~\citep{chen2022wavlm} proposes adopting the well-known HuBERT model~\citep{hubert} and training it with an additional denoising process. The authors apply a noising process to the training data and then predict the clean units from it. ContentVec~\citep{qian2022contentvec} is focused on the disentanglement of a speaker from self-supervised speech representation. The authors propose to use three disentanglement components. First, the student network is disentangled through two transformations. Then the representations are forwarded through a speaker condition component. Finally, voice-converted input data points are used to generate teacher labels.

\section{Conclusions}

In this work, we first propose a metric for evaluating the robustness of self-supervised speech representations applied for spoken language modeling tasks. Equipped with the aforementioned metric, we point out the lack of robustness in current state-of-the-art speech encoders with respect to simple signal variations that do not alter the spoken information. We then propose a simple and effective method to boost the robustness of the current approaches and demonstrate it on three state-of-the-art self-supervised speech representation models. We empirically show the efficacy of the proposed approach when considering encoding methods together with a textless speech-to-speech translation.

\clearpage

\section*{Broader Impact}
As for broader impacts, this work is the first (to the best of our knowledge) which analyzes self-supervised speech representation models, considering basic signal variations. We hope that with the provided analysis and evaluation, researchers working on spoken language modeling and self-supervised speech representation learning will consider reporting the proposed metric setup along with evaluation of down stream tasks.

\section*{Limitations}

The proposed method has several limitations that should be taken into consideration when employing it. First, the method relies on an existing model, e.g., k-means, which creates a dependency between the performance of the initial and the robust models. Second, the flow is not trained end-to-end, which can also limit its performance as end-to-end training allows improvement of the robustness of the whole representation. Lastly, to fully assess the effectiveness of the method, multiple metrics need to be examined. This can be a limitation as interpreting the results from multiple metrics may not be straightforward. However, it gives a more complete picture of the model's performance.

\bibliography{custom}
\bibliographystyle{acl_natbib}

\clearpage

\appendix
\section{Levenshtein Distance}\label{app:levenshtein}

Throughout the paper, we use a version of the Levenshtein distance. In this section, we detail the Levenshtein distance between two sequences. Let $x \in \{1,..,K\}^{T_x}$ and $y \in \{1,..,K\}^{T_y}$ be two discrete vectors, not necessary in the same size. Let us also denote the operator $\text{tail}(x)$ to return a copy of the vector $x$ without its first element. Then, the Levenshtein distance is defined recursively by $\text{Lev}(x, y) = $
\[
    \begin{cases}
      |x|, & \text{if}\ |y|=0 \\
      |y|, & \text{if}\ |x|=0 \\
      1 + \min \begin{cases}
                    \text{Lev}(\text{tail}(x), y) \\
                    \text{Lev}(x, \text{tail}(y)) \\ 
                    \text{Lev}(\text{tail}(x), \text{tail}(y)) \\
                \end{cases}, & \text{otherwise}
    \end{cases}\nonumber
\]
where $|x|, |y|$ are the lengths of the vectors $x$ and $y$ respectively. Note, in our implementation, we use deduplicated sequences.

% \section{Additional experimental details}\label{app:exp_details}

% As a quantizer, we use three non-linear layers. The dimensions of those layers are determined by the division floor of the difference between the upstream dimension (768 for the base and 1024 for the big model) to the number of units.
% % \begin{enumerate}
% %     \item 768,  (768 - 200) // 2
% %     \item (768 - 200) // 2, (768 - 200) // 4
% %     \item (768 - 200) // 4, 200
% % \end{enumerate}

\section{Extended Experimental Setup}\label{app:exp_setup}

\paragraph{Models.} We study our method using the base versions of HuBERT, wav2vec2, and WavLM. Similar to prior work, for HuBERT and WavLM, we use the ninth and sixth layers for wav2vec2. For readability, we report results for HuBERT in the main paper. The results for wav2vec2 and WavLM are presented in Appendix~\ref{app:results}. In our quantizer learning process, we use a learning rate of $0.0001$, a batch size of $32$, and Adam optimizer~\citep{adam}. Our quantizer is composed of three fully connected layers with LeakyReLU activation between them. The dimensions of those layers are determined by the division floor of the difference between the upstream dimension to the number of units. We train our quantizer using a single NVIDIA V100 GPU.

\paragraph{Datasets.} To match the current k-means popular training set, we use the Librispeech-100h to learn our quantizer~\citep{librispeech}. We analyze our metric using the `clean' and `other' development sets from Librispeech. The augmentations in all setups include time stretch, pitch shift, reverberation, and noise injection (exact parameters are detailed in Section~\ref{sec:aug}). For the sWUGGY and sBLIMP evaluations, we use the `big' transformer language model from~\citet{on_generative}.

This appendix begins with a detailed explanation on the Levenshtein distance (Section~\ref{app:levenshtein}). Then, in Section~\ref{app:results}, we present additional results. We report results on two additional state-of-the-art self-supervised speech representations. We show that our method is indeed effective for those representations as well as shown in the main paper.

\section{Additional Results}\label{app:results}
In the following, we provide additional results on the state-of-arts representations ``wav2vec2'' and ``WavLM''~\citep{baevski2020wav2vec, chen2022wavlm}.

Tables~\ref{table:wav2vec_ued} and~\ref{table:wavlm} present the UED scores for both the wav2vec2 and WavLM models. Using our method, we observe robustness improvements for both of the models. However, it is notable that the WavLM model is more robust than the wav2vec2 model. It is reasonable since the WavLM trained to be a more robust model using noisy training samples.

Tables~\ref{table:wavtovec_metric} and~\ref{table:wavlm_metric} present the discriminative and generative metrics for both wav2vec2 and WavLM. We observe a consistent improvement using our robust quantizer as in the robustness metrics. However, for the WavLM, the improvements are sometimes marginal (except for $k=50$ where k-means outperforms our method). The WavLM model is trained with a HuBERT architecture, with more data and noisy samples. Interestingly, while presenting better performance on various downstream tasks than HuBERT, their ABX, sWUGGY, and sBLIMP scores are lower.

\begin{table*}[t!]\centering
\begin{tabular}{llcccc}
    \toprule
    \multirow{2}[3]{*}{\# units} &\multirow{2}[3]{*}{Method} & \multicolumn{4}{c}{Augmentation} \\
    \cmidrule(lr){3-6}
     &                                    & Time                    & Pitch shift             & Reverberation           & Noise \\
    \midrule
    \multirow{3}{*}{50} &k-means          & 50.81\pmr{0.41}         & 58.66\pmr{1.16}         & 43.71\pmr{0.77}         &  32.17\pmr{0.61} \\
                        &Ours             & 38.74\pmr{0.45}         & 42.33\pmr{0.97}         & 33.69\pmr{0.73}         &  25.36\pmr{0.49} \\
                        &Ours (Iterative) & \textbf{36.68}\pmr{0.39}& \textbf{40.29}\pmr{1.04}& \textbf{33.28}\pmr{0.74}&  \textbf{23.99}\pmr{0.51} \\
    \midrule
    \multirow{3}{*}{100} &k-means         & 55.30\pmr{0.61}         & 65.23\pmr{0.91}         & 48.41\pmr{0.72}         & 33.97\pmr{0.46}\\
                         &Ours            & 42.32\pmr{0.46}         & 47.07\pmr{0.88}         & 36.83\pmr{0.71}         & 27.15\pmr{0.75}\\
                         &Ours (Iterative)& \textbf{40.43}\pmr{0.57}& \textbf{45.73}\pmr{0.90}& \textbf{36.34}\pmr{0.77}& \textbf{26.22}\pmr{0.59}\\
    \midrule
    \multirow{3}{*}{200} &k-means         & 59.85\pmr{0.39}         & 70.80\pmr{1.31}         & 53.13\pmr{0.67}         &  36.64\pmr{0.62}\\
                         &Ours            & 46.84\pmr{0.42}         & 51.60\pmr{1.21}         & \textbf{40.54}\pmr{0.66}&  32.61\pmr{0.67}\\
                         &Ours (Iterative)& \textbf{44.90}\pmr{0.35}& \textbf{49.59}\pmr{1.25}& \textbf{40.58}\pmr{0.62}&  \textbf{29.49} \pmr{0.57} \\
    \midrule
    \multirow{3}{*}{500} &k-means         & 66.12\pmr{0.48}         & 77.01\pmr{0.98}         & 59.69\pmr{1.01}         & 37.22\pmr{0.65}\\
                         &Ours            & 51.65\pmr{0.49}         & \textbf{55.40}\pmr{1.03}& 45.85\pmr{0.93}         & 33.17\pmr{0.62}\\
                         &Ours (Iterative)& \textbf{50.50}\pmr{0.53}& 57.12\pmr{1.02}         & \textbf{44.67}\pmr{0.98}& \textbf{31.92}\pmr{0.69}\\
    \bottomrule
\end{tabular}
\caption{Wav2vec2 unit edit distance}
\label{table:wav2vec_ued}
\end{table*}

\begin{table*}[t!]\centering
\begin{tabular}{llcccc}
    \toprule
    \multirow{2}[3]{*}{\# units} &\multirow{2}[3]{*}{Method} & \multicolumn{4}{c}{Augmentation} \\
    \cmidrule(lr){3-6}
     &                                    & Time                    & Pitch shift             & Reverberation           & Noise \\
    \midrule
    \multirow{3}{*}{50} &k-means          & 47.66\pmr{0.49}         & 52.93\pmr{1.02}         & 33.45\pmr{0.62}         &  28.46\pmr{0.61} \\
                        &Ours             & 39.12\pmr{0.43}         & 44.25\pmr{1.06}         & 31.58\pmr{0.62}         &  25.32\pmr{0.67} \\
                        &Ours (Iterative) & \textbf{36.79}\pmr{0.46}& \textbf{40.16}\pmr{1.05}& \textbf{25.73}\pmr{0.64}&  \textbf{25.01}\pmr{0.66} \\
    \midrule
    \multirow{3}{*}{100} &k-means         & 52.61\pmr{0.51}         & 58.44\pmr{0.72}         & 36.27\pmr{0.45}         & 29.44\pmr{0.64}\\
                         &Ours            & 43.55\pmr{0.53}         & 49.03\pmr{0.75}         & 30.54\pmr{0.44}         & 25.93\pmr{0.67}\\
                         &Ours (Iterative)& \textbf{42.11}\pmr{0.50}& \textbf{46.08}\pmr{0.74}& \textbf{28.88}\pmr{0.47}& \textbf{25.47}\pmr{0.59}\\
    \midrule
    \multirow{3}{*}{200} &k-means         & 58.50\pmr{0.42}         & 64.75\pmr{1.02}         & 41.05\pmr{0.54}         &  30.93\pmr{0.62}\\
                         &Ours            & 49.57\pmr{0.41}         & 53.48\pmr{1.09}         & 34.29\pmr{0.53}         &  26.66\pmr{0.65}\\
                         &Ours (Iterative)& \textbf{47.82}\pmr{0.46}& \textbf{52.47}\pmr{1.01}& \textbf{32.88}\pmr{0.55}&  \textbf{26.09} \pmr{0.62} \\
    \midrule
    \multirow{3}{*}{500} &k-means         & 64.25\pmr{0.67}         & 70.55\pmr{0.75}         & 45.63\pmr{0.83}         &  33.17\pmr{0.71}\\
                         &Ours            & 55.41\pmr{0.64}         & 59.79\pmr{0.87}         & 42.85\pmr{0.78}         &  28.46\pmr{0.79}\\
                         &Ours (Iterative)& \textbf{52.92}\pmr{0.69}&\textbf{57.84}0\pmr{0.81}& \textbf{40.46}\pmr{0.81}&  \textbf{27.09}\pmr{0.72}\\
    \bottomrule
\end{tabular}
\caption{WavLM unit edit distance}
\label{table:wavlm}
\end{table*}

\begin{table*}[t!]\centering
\begin{tabular}{llccccccc}
    \toprule
    \multirow{2}[3]{*}{\# units} &\multirow{2}[3]{*}{Method} & \multicolumn{2}{c}{ABX (clean) $\downarrow$} & \multicolumn{2}{c}{ABX (other)$\downarrow$}& \multirow{2}[3]{*}{sWUGGY $\uparrow$} & \multirow{2}[3]{*}{sBLIMP $\uparrow$} \\
    \cmidrule(lr){3-4}\cmidrule(lr){5-6}
     &&Within & Across & Within & Across&& \\
    \midrule
    \multirow{3}{*}{50} &k-means& 12.03&15.31&13.61&19.07&\textbf{49.76}&53.92\\
    &Ours& 11.18&13.82&13.34&18.39&-&-\\
    &Ours (Iterative)& \textbf{10.35}&\textbf{12.75}&\textbf{12.64}&\textbf{17.29}&49.65&\textbf{55.29}\\
    \midrule
    \multirow{3}{*}{100} &k-means& 11.27&13.99&13.06&17.11&51.63&53.87\\
    &Ours& 9.86&11.81&11.44&16.63&-& \\
    &Ours (Iterative)& \textbf{9.24}&\textbf{11.30}&\textbf{11.37}&\textbf{16.14}&\textbf{51.90}&\textbf{54.95}\\
    \midrule
    \multirow{3}{*}{200} &k-means& 11.13&14.42&12.37&18.02&51.29&54.99\\
    &Ours& 10.19&12.41&11.85&17.52&-&-\\
    &Ours (Iterative)&\textbf{9.00}&\textbf{11.11}&\textbf{11.49}&\textbf{16.53}&\textbf{51.99}&\textbf{55.67}\\
    \midrule
    \multirow{3}{*}{500} &k-means& 12.06&15.61&13.77&19.94&52.21&54.32\\
    &Ours&10.76&13.83&13.52&19.60&-&-\\
    &Ours (Iterative)&\textbf{10.16}&\textbf{12.42}&\textbf{12.56}&\textbf{18.24} &\textbf{52.93}&\textbf{55.17}\\
    \bottomrule
\end{tabular}
\caption{Wav2vec2 discriminative and generative evaluation metrics.}
\label{table:wavtovec_metric}
\end{table*}

\begin{table*}[t!]\centering
\begin{tabular}{llccccccc}
    \toprule
    \multirow{2}[3]{*}{\# units} &\multirow{2}[3]{*}{Method} & \multicolumn{2}{c}{ABX (clean) $\downarrow$} & \multicolumn{2}{c}{ABX (other)$\downarrow$}& \multirow{2}[3]{*}{sWUGGY $\uparrow$} & \multirow{2}[3]{*}{sBLIMP $\uparrow$} \\
    \cmidrule(lr){3-4}\cmidrule(lr){5-6}
     &&Within & Across & Within & Across&& \\
    \midrule
    \multirow{3}{*}{50} &k-means& 7.60 & 9.06 & \textbf{9.22} & 12.99 & 63.91& 55.29\\
    &Ours& 7.41 & 8.68 & 9.51 & \textbf{11.78} &-&-\\
    &Ours (Iterative)& \textbf{7.19} & \textbf{8.25} & 9.41 & 11.87 & \textbf{64.87}&\textbf{55.81}\\
    \midrule
    \multirow{3}{*}{100} &k-means& 6.91 & 8.06 & 8.95 & 11.86 &63.61&54.59\\
    &Ours& \textbf{6.02} & 7.13 & 8.36 & \textbf{10.95}&-&-\\
    &Ours (Iterative)& 6.39 & \textbf{7.02} & \textbf{8.17} & 11.21&\textbf{63.99}&\textbf{54.97}\\
    \midrule
    \multirow{3}{*}{200} &k-means& 6.74 & 8.12 & 8.76 & 12.09 &65.97&55.59\\
    &Ours& \textbf{6.40} & \textbf{7.45} & \textbf{8.61} & \textbf{11.49} &-&-\\
    &Ours (Iterative)& 6.51 & 7.73 & 8.93 & 11.94 &\textbf{66.90}&\textbf{55.89}\\
    \midrule
    \multirow{3}{*}{500} &k-means& 7.14 & 8.10 & 9.09 & 11.70 &64.56&55.91\\
    &Ours& \textbf{7.03} & \textbf{7.91} & \textbf{8.99} & \textbf{11.21} &-&-\\
    &Ours (Iterative)& 7.08 & 7.87 & 9.03 & 11.54 &\textbf{65.81}&\textbf{56.09}\\
    \bottomrule
\end{tabular}
\caption{WavLM discriminative and generative evaluation metrics.}
\label{table:wavlm_metric}
\end{table*}

\end{document}